\documentclass[letterpaper]{article} 
\usepackage{aaai24}  
\usepackage{times}  
\usepackage{helvet}  
\usepackage{courier}  
\usepackage[hyphens]{url}  
\usepackage{graphicx} 
\usepackage{natbib}  
\usepackage{caption} 
\usepackage{algorithm}
\usepackage{algorithmic}
\usepackage[namelimits]{amsmath} 
\usepackage{amssymb}
\usepackage{amsfonts}
\usepackage{mathrsfs}
\usepackage{amsmath}
\usepackage{multirow}
\usepackage{booktabs}
\usepackage{subcaption}
\usepackage{bbding}

\usepackage{newfloat}
\usepackage{listings}
\DeclareCaptionStyle{ruled}{labelfont=normalfont,labelsep=colon,strut=off} 
\floatstyle{ruled}
\newfloat{listing}{tb}{lst}{}
\floatname{listing}{Listing}
\title{Style$^2$Talker: High-Resolution Talking Head Generation \\ with Emotion Style and Art Style}
\author{
	Shuai Tan,
	Bin Ji,
	Ye Pan\thanks{Corresponding author.}
}
\affiliations{
	Shanghai Jiao Tong University \\

	\{tanshuai0219, bin.ji, whitneypanye\}@sjtu.edu.cn
}

\usepackage{bibentry}

\begin{document}
	
	\maketitle
	
	\begin{abstract}
		Although automatically animating audio-driven talking heads has recently received growing interest, previous efforts have mainly concentrated on achieving lip synchronization with the audio, neglecting two crucial elements for generating expressive videos: emotion style and art style. In this paper, we present an innovative audio-driven talking face generation method called Style$^2$Talker. It involves two stylized stages, namely Style-E and Style-A, which integrate text-controlled emotion style and picture-controlled art style into the final output. In order to prepare the scarce emotional text descriptions corresponding to the videos, we propose a labor-free paradigm that employs large-scale pretrained models to automatically annotate emotional text labels for existing audio-visual datasets. Incorporating the synthetic emotion texts, the Style-E stage utilizes a large-scale CLIP model to extract emotion representations, which are combined with the audio, serving as the condition for an efficient latent diffusion model designed to produce emotional motion coefficients of a 3DMM model. Moving on to the Style-A stage, we develop a coefficient-driven motion generator and an art-specific style path embedded in the well-known StyleGAN. This allows us to synthesize high-resolution artistically stylized talking head videos using the generated emotional motion coefficients and an art style source picture. Moreover, to better preserve image details and avoid artifacts, we provide StyleGAN with the multi-scale content features extracted from the identity image and refine its intermediate feature maps by the designed content encoder and refinement network, respectively. Extensive experimental results demonstrate our method outperforms existing state-of-the-art methods in terms of audio-lip synchronization and performance of both emotion style and art style.
	\end{abstract}
	
	\section{Introduction}
	The automatic animation of images plays a crucial role in computer graphics and vision, finding applications in various fields such as film production, virtual avatars, and social media~\cite{pataranutaporn2021ai}. To enable more extensive applications, two essential elements come into play: emotion style and art style, which we refer to as style$^2$. Emotion style allows users to convey communicative information more efficiently with diverse expressions~\cite{PaulEkman2005WhatTF}, while art style can evoke different human experiences, leading to stronger visual effect and applications in entertainment~\cite{yang2022vtoonify}. However, the generation of audio-driven talking head videos with both styles from a regular face photo has not been extensively explored.
	
	\begin{figure}[t]
		\centering
		\includegraphics[width=0.94\linewidth]{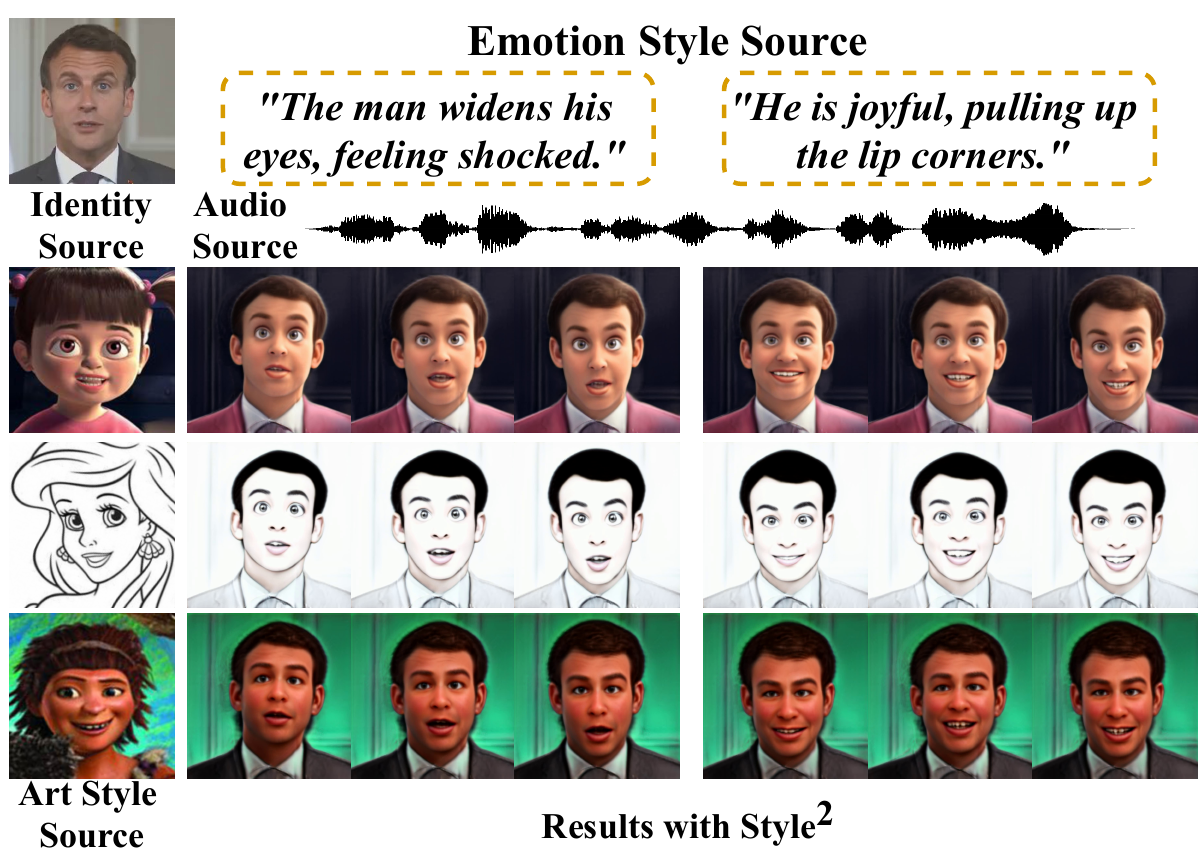}
		\caption{Illustrative animations produced by our Style$^2$Talker. Our approach takes an identity image and an audio clip as inputs and generates a talking head with emotion style and art style, which are controlled respectively by an emotion source text and an art source picture.}
		\label{fig:teaser}
	\end{figure}
	
	When it comes to emotion style, previous works have either used a one-hot emotion label as the emotion source~\cite{ji2021audio, SanjanaSinha2022EmotionControllableGT}, limiting the range of expressions, or relied on an additional emotion video~\cite{ji2022eamm}, which can be inconvenient as finding a video with the desired emotion style might not always be feasible. In contrast, we design our system to enable more user-friendly input by allowing users to provide a text description of the emotion style, encompassing emotion categories and detailed facial muscle movements. On the other hand, assuming that using a picture is more suitable and visual to indicate art style like the rendering color of specific part, face shape, etc, we prefer an art picture for the art style reference. While there have been several efforts in single image style transfer~\cite{yang2022pastiche, choi2020stargan}, these methods face challenges when generating continuous videos driven by an audio clip. As a result, the objective of our study is to develop a system capable of synthesizing high-resolution talking face videos, whose identity, mouth shapes, emotion style and art style align with the input identity image, audio clip, emotion textual description and art picture, respectively. 
	
	
	Specifically, we present a novel framework named Style$^2$Talker, designed to achieve this objective by involving emotionally stylized stage Style-E and artistically stylized stage Style-A. 
	We leverage 3DMM (3D Morphable Model)~\cite{deng2019accurate} coefficients as a brief intermediate representation to bridge the gap between the two stylized stages. The Style-E stage develops a latent diffusion model that acts as the emotionally stylized motion generator. By taking the text as emotion style descriptors and audio as motion driving sources, the generator produces high-quality, realistic expression coefficient sequences that convey the desired emotion styles. The motivations for using diffusion model are: (1) Talking face generation with text-driven emotion style is a classic conditional generation task, and diffusion model exhibits excellent performance in this area~\cite{stypulkowski2023diffused} and is more stable than conditional GAN~\cite{mirza2014conditional} for training by removing adversarial process. (2) The diffusion and denoising process in the training phase allows the model to be more robust and precise in mining the expected results from the random noise based on the provided conditions during inference. However, training such a text-driven emotional generation model necessitates textual descriptions for emotional expression, which is absent until now. To address this, we devise an automatically annotated method by leveraging large-scale pretrained models. With the generated text descriptors, we incorporate the CLIP text encoder~\cite{radford2021learning} and an audio encoder to extract emotion representations and audio features, which are passed through the motion generator as denoising conditions for emotion stylization. To optimize the inference time, we employ a simpler and more efficient diffusion model as our motion generator.

	The Style-A stage is based on DualStyleGAN~\cite{yang2022pastiche}, which introduces an art-specific path compared to StyleGAN~\cite{karras2020analyzing} to transfer the art style of a single image to those of an art reference picture. To generate continuous artistic frames that align with 3DMM coefficients from the Style-E stage, we draw inspiration from the finding that the spatial feature map in the $\mathcal{F}$ space of StyleGAN is highly related to expressing the pose of the generated images~\cite{yin2022styleheat}. Therefore, we employ a coefficient-driven motion generator to produce flow maps from coefficients, which are then used to warp the spatial feature map. This way, StyleGAN is able to synthesize talking face videos with the aid of the continuously warped spatial feature maps. However, the current results suffer from details loss due to the GAN Inversion encoder of DualStyleGAN, which primarily focuses on face reconstruction and neglects preserving detailed information like background and texture. To tackle this issue, we introduce a content encoder that provides StyleGAN with additional multi-level content features extracted from the identity image. These features supplement the texture details using skip connections, following~\citet{yang2022vtoonify}. Furthermore, we design a refinement network to eliminate potential ghost shadows caused by the misalignment between the warped spatial feature maps and fixed content features. Overall, the Style$^2$Talker framework offers a promising approach to achieve emotionally and artistically stylized talking face generation with improved continuity and realism. Through extensive experiments, we demonstrate the effectiveness and superiority of our method over state-of-the-art approaches.
	
	Our principal contributions are summarized as follows:
	\begin{itemize}
		\item We present a novel system that facilitates high-resolution talking head generation with emotion style and art style. To the best of our knowledge, we are the first to combine both styles in the context of talking-head task under the guidance of emotion text and art picture simultaneously.
		\item We explore an innovative labor-free method for automatically generating text descriptions to serve as emotion style sources with the assistance of large-scale pretrained models. By incorporating audio clips and synthetic text, we introduce an efficient diffusion model to synthesize emotionally stylized motion coefficients.
		\item We demonstrate a successful extensive application of modified StyleGAN to enable high-resolution talking head generation with the desired art style driven by the generated emotional coefficients and an art picture.
	\end{itemize}
	
	\begin{figure*}[t]
		\centering
		\includegraphics[width=1\linewidth]{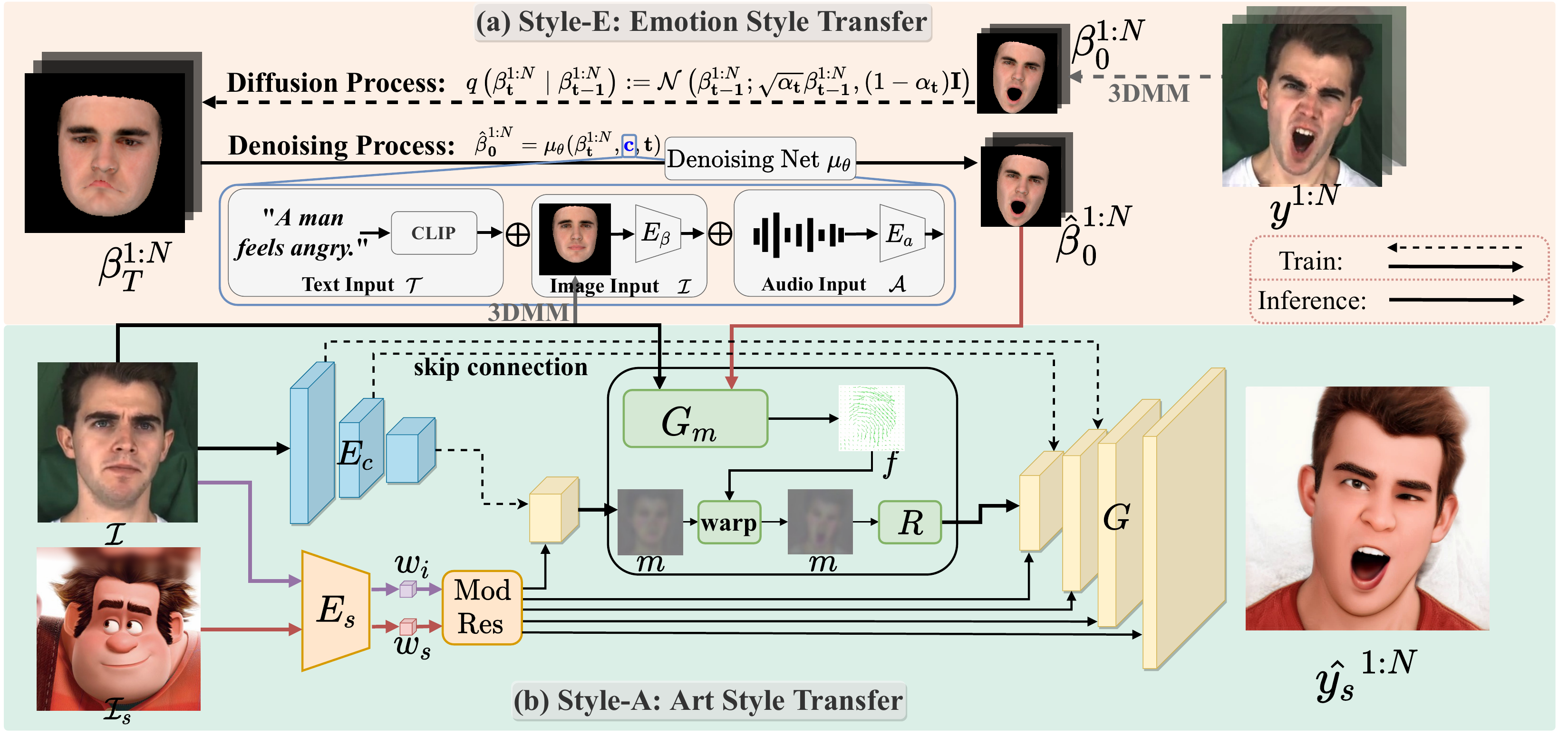}
		\caption{The overview of the proposed $\text{Style}^2\text{Talker}$. (a) Style-E: Emotion Style Transfer. In the diffusion process, we start by extracting the 3D expression coefficient sequence $\mathbf{\beta}^{1:N}_{0}$ from the ground truth video $y^{1:N}$. Then, we iteratively add Gaussian noise by $q\left(\mathbf{\beta}^{1:N}_\mathbf{t} \mid \mathbf{\beta}^{1:N}_{\mathbf{t-1}}\right) :=\mathcal{N}\left(\mathbf{\beta}^{1:N}_{\mathbf{t-1}} ; \sqrt{\alpha_\mathbf{t}} \mathbf{\beta}^{1:N}_{\mathbf{t-1}}, (1-\alpha_\mathbf{t}) \mathbf{I}\right)$. A simple MLP-based denoising network  $\mu_\theta$ is trained to denoise the noisy parameters $\mathbf{\beta}_\mathbf{t}$ at time $\mathbf{t}$ based on the conditioning signal $\mathbf{c}$, and $\mathbf{c}$ comprises text $\mathcal{T}$, identity image $\mathcal{I}$ and audio $\mathcal{A}$. (b) Style-A: Art Style Transfer. We employ a pre-trained encoder $E_s$ from pSp~\cite{richardson2021encoding} to embed the identity image $\mathcal{I}$ and art source image $\mathcal{I}_s$ to the latent code $w_i$ and $w_s$, which are fed into a ModRes to merge the style code. To alleviate content loss caused by the pSp~\cite{richardson2021encoding}, we introduce another Content Encoder $E_c$ to extract multi-level content features. These features are then fed into StyleGAN $G$ through skip connections and combined with style codes, serving as the input of $G$. To enable continuous frames generation, the produced emotional coefficients $\hat{\mathbf{\beta}}^{1:N}_{0}$ are converted to flow field map $f$, which in turn warps feature map $m$ of StyleGAN to $\hat{m}$, achieving talking head generation with style$^2$.}
		\label{fig:overview}
	\end{figure*}
	
	\section{Related Work}
	\subsection{Audio-driven Talking Head Synthesis}
	In recent times, audio-driven face animation has garnered significant attention owing to its wide range of applications in fields such as like film-making and virtual reality. Existing approaches can be broadly categorized into two groups: person-specific methods and person-independent methods. Person-specific methods~\cite{guo2021ad,ye2023geneface} focus on training models for specific individuals using videos featuring those individuals. By avoiding the inconsistency of speech styles across different speakers~\cite{wang2022one}, person-specific methods achieve superior performance on specific individual. However, a drawback of person-specific methods is their limited applicability to other identities. On the contrary, person-independent models generalize their capabilities to arbitrary identity by training on multi-speaker audio-visual datasets. These methods not only achieve lip motions synchronized with the audio by animating the face/mouth regions~\cite{alghamdi2022talking, prajwal2020lip}, but also generate head motions that align with the rhythm of the audio to make the outputs more realistic~\cite{zhang2022sadtalker}.

	Recent advancements~\cite{ji2021audio, pan2023emotional,tan2023emmn, tan2024saas,pan2024expressive} have also explored the synthesis of emotional expressions in talking faces. \citet{ji2021audio} and \citet{SanjanaSinha2022EmotionControllableGT} utilize one-hot emotion labels as input to generate emotional talking faces, while others~\cite{ji2022eamm, ma2023styletalk} resort to another video for emotion source. In contrast, our approach offers a more user-friendly control by allowing users to input easy-to-use text descriptions to suggest the desired emotion style. In this way, users have a straightforward and intuitive way to specify the emotions they want to see in the generated talking faces. Furthermore, we adopt a modified version of StyleGAN~\cite{karras2020analyzing} to generate high-resolution talking face videos with the art style given an art reference image. This modification overcomes the issues of blurriness and low-resolution outputs encountered in existing methods. 
	
	\subsection{Diffusion Generative Models}
	The field of Diffusion Generative Models~\cite{Dhariwal_Nichol_2021,Ho_Jain_Abbeel_2020, Nichol_Dhariwal_2021} has witnessed a remarkable surge in advancements, showcasing impressive performance in conditional generation tasks~\cite{Dhariwal_Nichol_2021}. Talking face generation has also seen extensions of diffusion models in several works~\cite{stypulkowski2023diffused}, where a general paradigm involves training U-net-like denoising network conditioned on time step and audio. However, a notable limitation of diffusion models lies in their extensive inference time, mainly due to the computationally intensive image-level denoising network and the involvement of multiple denoising processes. In light of this challenge, we extract the 3DMM coefficients from videos as brief intermediate representation and introduce a simpler and more efficient MLP-based denoising network~\cite{du2023avatars}, which significantly reduce the inference time.

	\subsection{Face Manipulation by StyleGAN}
	StyleGAN~\cite{karras2019style, karras2020analyzing} has successfully proved its power to generate impressively realistic and high-resolution images. Various valuable modifications on StyleGAN have been explored for interesting applications, which fall into two main categories. The first category focuses on editing input face attributes, such as gender, pose, or age, within the latent space of StyleGAN~\cite{richardson2021encoding, abdal2019image2stylegan}. Inspired by these advancements,~\citet{alghamdi2022talking} propose editing the latent space conditioned on audio, which enables the generation of talking head videos. On the other hand, the second category deals with editing the art style. Studies~\cite{yang2022pastiche, yang2022vtoonify} focus on transferring art style from an original image to a given style reference, often employing an extrinsic style path. However, their emphasis has primarily been on style transfer for individual frames. In contrast, our approach goes beyond these tasks by combining both attribute control and art style transfer to produce audio-driven artistically stylized talking head videos—an interesting but unexplored field.

	\section{Proposed Method}

	Figure~\ref{fig:overview} illustrates the pipeline of our $\text{Style}^2\text{Talker}$, which is composed of two stylized stages. In the Style-E stage, we leverage the diffusion model framework, incorporating both the diffusion and denoising processes. During the denoising process, the denoising network iteratively denoises the sampled random noise vector, generating an emotionally stylized coefficient sequence of the 3DMM model conditioned on the input text, image, and audio. In the Style-A stage, we introduce an elaborately modified StyleGAN to stylize the input face from the original art style to the given reference one. In subsequent sections, we will provide a detailed explanation of each stage within our proposed framework.
	
	\subsection{Data Preparation}
	To enhance user control flexibility, we enable text-guidance emotion style generation, wherein users can describe their desired emotion style using text. To realize this capability, a text-emotion audio-visual dataset is required. In comparison to the existing emotional audio-visual dataset like MEAD~\cite{wang2020mead}, the text-emotion audio-visual dataset additionally contains accurate text descriptions of emotions performed in the videos. Regrettably, such an open-source dataset does not currently exist. 
	
	While MEAD provides 8 general emotion descriptions, they may not be sufficient to accurately express the emotion style. Therefore, we devise a labor-free dataset generation pipeline that leverages large-scale pretrained models to extend MEAD with more detailed text descriptions of the emotion style, including semantic-rich adjectives and detailed facial motion descriptions represented by facial action units (AUs)~\cite{ekman1978facial} following~\cite{hong2020face}. We strongly recommend reading the detailed flowchart in the supplementary materials. Each AU corresponds to specific movements in different parts of the face, and combinations of multiple AUs describe the overall emotion style. Particularly, we employ OpenFace~\cite{baltrusaitis2018openface} to detect continuous intensity for AUs, where we set an intensity threshold to determine AU activated/inactivated, and activated AUs are further divided into three discrete intensities as level labels. Subsequently, we turn to GPT-3~\cite{brown2020language}, a pretrained language model that encodes real-world knowledge, for providing more synonymous annotations for emotion classes and AUs with different levels. With the obtained textual descriptions of AU and emotion style, GPT-3 has the flexibility to generate several candidate sentences with varied syntaxes that provide comprehensive text descriptions of emotion styles for the videos in MEAD. 
	
	To filter out noisy samples from the generated emotion style text, we implement a data post-cleaning strategy by assessing the similarity between emotional videos in the MEAD dataset and each candidate textual description. The large-scale pretrained model CLIP~\cite{radford2021learning} is well-suited for this task as its training strategy, where closer text descriptions to videos result in higher similarity. Therefore, we employ the text encoder and image encoder of CLIP to extract corresponding features from both the texts and frames of each video. The cosine similarity between the text feature from each candidate sentence and the image feature is calculated and ranked. Consequently, the five candidate sentences with the highest similarity for each video are retained as the final text descriptions. During training, we randomly select one sentence from these five candidates as the textual input for each iteration, ensuring a diverse and robust training process.

	\subsection{Style-E: Emotion Style Transfer}
	In this stage, we adopt the diffusion model~\cite{song2020denoising} for producing stylized coefficient sequences from multiple inputs, given its impressive performance in conditional generation tasks. Starting with a stylized video $y^{1:N}_0$ containing $N$ frames, we extract 3DMM~\cite{blanz1999morphable} expression coefficients $\beta^{1:N}_0$\footnote{Please note that $\beta$ in our paper refers to expression coefficients of 3DMM, instead of hyper-parameter in original DDPM paper.} using 3D reconstruction method~\cite{deng2019accurate}. The forward diffusion process is formulated as follows:
	\begin{equation}
		q\left(\mathbf{\beta}^{1:N}_\mathbf{t} \mid \mathbf{\beta}^{1:N}_{\mathbf{t}-1}\right) :=\mathcal{N}\left(\mathbf{\beta}^{1:N}_{\mathbf{t}-1} ; \sqrt{\alpha_t} \mathbf{\beta}^{1:N}_{\mathbf{t}-1}, (1-\alpha_\mathbf{t}) \mathbf{I}\right),
	\end{equation}
	where $\mathbf{t} \in [1,...,T]$ represents the step number of diffusion and $\alpha_\mathbf{t}$ denotes the noise schedule at $\mathbf{t}$-step.

	During the denoising process, we train a denoising network $\mu_\theta$ to generate an emotionally stylized motion sequence from random noise $\mathbf{\beta}^{1:N}_{T} \sim \mathcal{N}(0,\mathbf{I})$. Intrigued by the success of diffusion prior in DALL$\cdot$E2~\cite{ramesh2022hierarchical}, $\mu_\theta$ directly predicts emotional motion sequence $\hat{\mathbf{\beta}^{1:N}_{\mathbf{0}}} = \mu_\theta(\mathbf{\beta}^{1:N}_{\mathbf{t}},\mathbf{c},\mathbf{t})$, instead of predicting the noise as in the vanilla DDPM~\cite{Ho_Jain_Abbeel_2020}. In our context, the condition $\mathbf{c}=\mathcal{T} \oplus \mathcal{I} \oplus \mathcal{A}$ concatenates multiple features extracted from a textual description $\mathcal{T}$ of emotion style, an identity image $\mathcal{I}$ and an audio clip $\mathcal{A}$ via corresponding encoders. The objective function is then formulated as:
	\begin{equation}
		L_\text{style1} = \mathbb{E}_{\beta_0^{1: N} \sim q\left(\beta_0^{1: N}\right), \mathbf{t} \sim [1:T]}\left[\left\|\beta_0^{1: N}-\hat{\beta}_0^{1: N}\right\|_2^2\right].
	\end{equation}
	
	To address the inherent limitation of slow inference time in the diffusion model~\cite{Ho_Jain_Abbeel_2020}, we implement $\mu_\theta$ with a more lightweight and efficient MLP-based architecture following~\citet{du2023avatars}. Furthermore, we leverage the DDIM~\cite{song2020denoising} technique, which allows us to sample only 5 steps instead of 1000 during inference, which contributes to a substantial decrease in inference time.
	
	\begin{table*}[t]
		\centering
		\resizebox{\linewidth}{!}{
			\begin{tabular}{l|ccccc|ccccc}
				\toprule
				\multicolumn{1}{c}{\multirow{2}[4]{*}{\textbf{Method}}} & \multicolumn{5}{c}{\textbf{MEAD}~\cite{wang2020mead}} & \multicolumn{5}{c}{\textbf{HDTF}~\cite{zhang2021flow}}\\
				\cmidrule(lr){2-6}  \cmidrule(lr){7-11}  \multicolumn{1}{c}{} & \multicolumn{1}{c}{SSIM$\uparrow$} & \multicolumn{1}{c}{FID$\downarrow$} & \multicolumn{1}{c}{M-LMD$\downarrow$} & \multicolumn{1}{c}{F-LMD$\downarrow$} & \multicolumn{1}{c}{$\text{Sync}_\text{conf}\uparrow$} & \multicolumn{1}{c}{SSIM$\uparrow$} & \multicolumn{1}{c}{FID$\downarrow$} & \multicolumn{1}{c}{M-LMD$\downarrow$} & \multicolumn{1}{c}{F-LMD$\downarrow$} & \multicolumn{1}{c}{$\text{Sync}_\text{conf}\uparrow$}
				\\
				
				\midrule
				\multirow{1}[2]{*}{VT+MakeItTalk} & 0.692 &82.577     &6.696     &  5.948   &  0.734   &  0.630   & 50.009    &   6.907  & 6.279    & 0.857  \\
				\multirow{1}[2]{*}{VT+Wav2Lip} & 0.700 & 129.893   &6.153     &  5.465  &  \textbf{3.663 }   &  0.656  &48.888    &   6.279  &5.662      & 2.479  \\
				\multirow{1}[2]{*}{VT+Audio2Head} & 0.660 & 75.253     &9.032     & 9.856  &  1.772   &   0.588   & 48.91    &  7.087   &6.930      &2.559  \\
				\multirow{1}[2]{*}{VT+PC-AVS} & 0.624 & 148.015    &14.250     &  12.758   &  2.664    &   0.436   & 110.108   &   9.678   &13.321    &2.545  \\
				
				\multirow{1}[2]{*}{VT+AVCT} & 0.632 & 63.222    &12.461    & 11.355   &  2.841    &   0.583   &44.616    &  12.005  & 10.739      & \textbf{3.515} \\
				
				\midrule
				
				\multirow{1}[2]{*}{VT+EAMM} & 0.690 & 73.167     &6.541    &  6.247   &1.801   &  0.555   & 65.048    & 7.771  &7.872    &2.355\\
				
				\multirow{1}[2]{*}{VT+StyleTalk} &0.726 & 91.661   &4.343    &  4.696  &1.987   &  0.641  & 50.974   &  6.268 &6.404     & 2.461\\
				\textbf{Style$^2$Talker} & \textbf{0.795} & \textbf{23.207}     &\textbf{3.317}     &   \textbf{2.696}   &  2.847 &   \textbf{0.718}   & \textbf{23.330}   &   \textbf{3.046}   &\textbf{2.791}     & 2.734  \\ 
				\midrule
				\multirow{1}[2]{*}{GT+VT} & 1.000 & 0.000     &0.000   &   0.000  & 2.985   &   1.000  & 0.000     &0.000   &   0.000     &2.262 \\
				\bottomrule
			\end{tabular}%
		}
		\caption{Quantitative comparisons with state-of-the-art methods. We evaluate each method on MEAD and HDTF datasets. For assessing the art style, we apply VToonofy to stylize the input frame, denoted as VT'. `Style-E' and `Style-A' refers to emotion and art style. The symbols `$\uparrow$' and `$\downarrow$' indicate higher and lower metric values for better results, respectively.}
		\label{tab:quantitative}%
	\end{table*}%
	
	\begin{figure*}[t]
		\centering
		\includegraphics[width=0.95\linewidth]{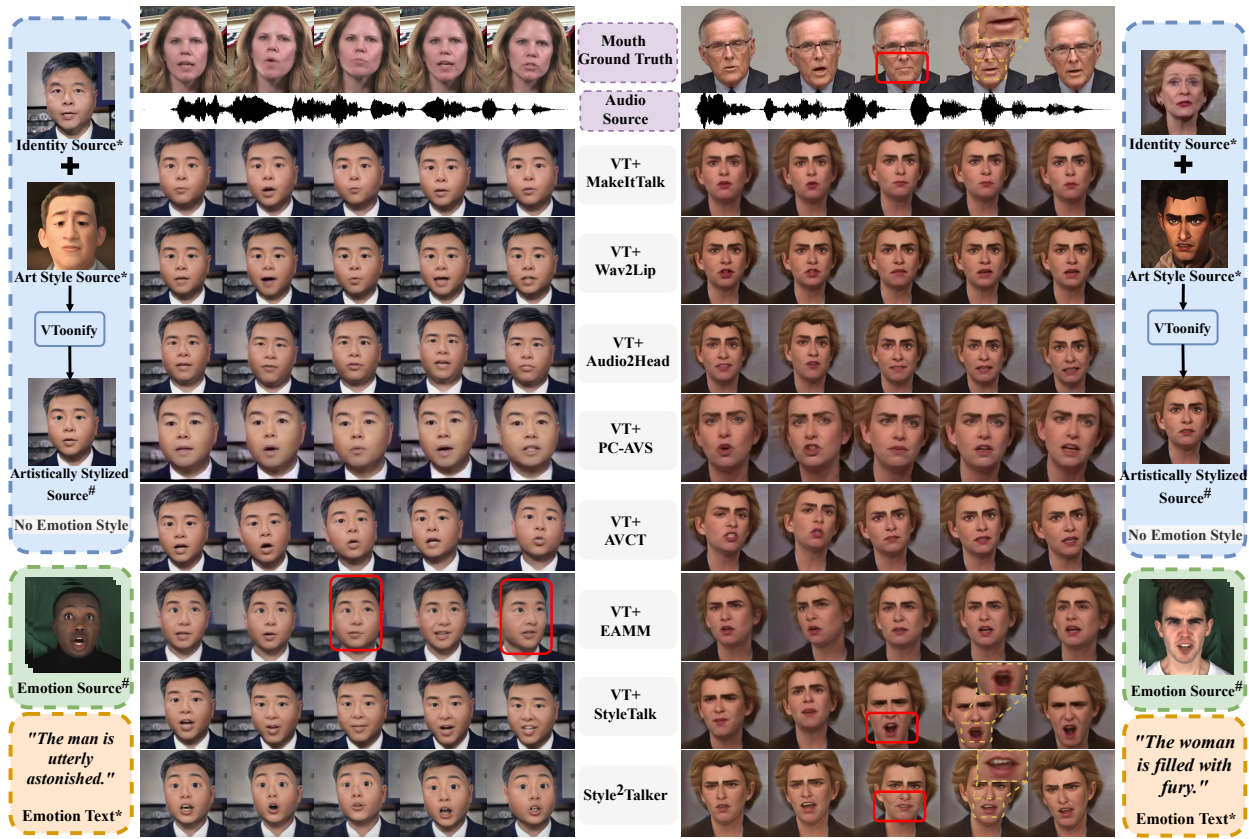}
		\caption{Qualitative comparisons with state-of-the-art methods. The input for our Style$^2$Talker and SOTAs are marked by `*' and `$\#$', respectively. We keep the original color style for better comparison.}
		\label{fig:compare}
	\end{figure*}
	
	\subsection{Style-A: Art Style Transfer}
	To perform high-resolution talking heads with art style, we build our framework upon DualStyleGAN~\cite{yang2022pastiche}, which introduces a new extrinsic art style path and ModRes block in comparison to vanilla StyleGAN~\cite{karras2020analyzing} to artistically stylize a single image. Specifically, given an identity image $\mathcal{I}$ and an art style reference image $\mathcal{I}_s$, a GAN Inversion encoder $E_s$~\cite{richardson2021encoding} is employed to encode them into latent code $w_i$, $w_s$, which provide content information and art style information, respectively. To effectively merge such two information and preserve the generative space and behavior of the pre-trained StyleGAN $G$ simultaneously, the ModRes block is introduced to adjust the structure styles of $w_i$ based on $w_s$ in a residual manner, enabling the pre-trained StyleGAN $G$ to transfer the art style of the identity image $\mathcal{I}$ into that of $\mathcal{I}_s$ with high resolution. However, despite its success in art style transfer, this approach faces challenges when generating continuous talking head videos under the guidance of the predicted motion sequence $\hat{\beta}_0^{1: N}$. Additionally, the stylized images suffer from image detail loss due to the performance limitation of GAN Inversion~\cite{wang2022high}.

	Sparked by the observation~\cite{yin2022styleheat} that spatial feature map in $64\times64$ layer reflects the pose and expression of the generated image, we introduce a motion generator $G_m$ to produce a flow field $f$ from the spatial feature map $m$ to the desired feature map $\hat{m}$. The motion generator $G_m$ consists of an image encoder, a flow decoder and a coefficient encoder. Specifically, for each $\hat{\beta}^{t}_0 \in \hat{\beta}^{1:N}_0$, the image encoder encodes identity image $\mathcal{I}$ into multiple feature maps $x_i$ using convolutional layers. These feature maps are then passed through an adaptive instance normalization (AdaIN~\cite{Huang_Belongie_2017}) operations as: $\operatorname{AdaIN}\left(\mathbf{x}_i, \mathbf{y}\right)=\mathbf{y}_{s, i} \frac{\mathbf{x}_i-\nu\left(\mathbf{x}_i\right)}{\sigma\left(\mathbf{x}_i\right)}+\mathbf{y}_{b, i}$, where $\nu(\cdot)$ and $\sigma(\cdot)$ represent the average and variance operations. $\mathbf{y} = (\mathbf{y}_{s}, \mathbf{y}_{b})$ are generated from the predicted 3DMM coefficients $\hat{\beta}^{t}_0$ through the coefficient encoder. By incorporating the flow decoder, we obtain the flow field $f$ and use it to warp the spatial feature map $m$ to $\hat{m}$, whose expressions and poses align with predicted emotion coefficients. 
	
	As for the second problem above, we adopt a Content Encoder $E_c$~\cite{yang2022vtoonify} to obtain multi-scale content features, which are passed to the StyleGAN $G$ to supplement the texture details via skip connections. By including additional multi-scale identity features, we can effectively preserve the image details of the original frame compared to previous StyleGAN-based methods that solely rely on the style condition, enabling high-fidelity artistically stylized images. Please note that since the skip connection passes texture information to the layer behind $\hat{m}$ as shown by the dotted line in Figure~\ref{fig:overview}, and the texture information is extracted from the original image and is not aligned with $\hat{m}$, ghosts will inevitably appear in the output. To this end, we construct a Refinement Network $R$ to adaptively tune $\hat{m}$ to rectify these artifacts. With the refined spatial feature maps, $G$ subsequently generates a continuous sequence of stylized frames $\hat{y_s}^{1:N}$, achieving talking head generation with style$^2$.
	
	During Style-A training, we freeze the weights of $E_s$ and $G$ which are pretrained in DualStyleGAN, and optimize the remaining networks (i.e., $E_c$, $G_m$ and $R$). To obtain the ground truth for training, we employ VToonify~\cite{yang2022vtoonify} to produce artistically stylized frames $y_s^{1:N} = \operatorname{VToonify}(y^{1:N}, I_s)$ from video $y^{1:N}$. Concretely, we import reconstruction loss $L_\text{rec}$ and perceptual loss $L_\text{prec}$~\cite{Johnson_Alahi_Fei-Fei_2016} to constrain the networks.
	\begin{equation}
		\label{eq:stage2}
		L_\text{style2} =\underbrace{\left\|y_s^{t}-\hat {y_s^{t}}\right\|_2}_{L_\text{rec}}+\underbrace{\lambda\left\|\text{VGG}(y_s^{t})-\text{VGG}(\hat {y_s^{t}})\right\|_1}_{L_\text{prec}},
	\end{equation}
	where $\text{VGG}$ implies the pretrained VGG network~\cite{Simonyan_Zisserman_2015}, and $\lambda = 0.1$ refers to the weight of $L_\text{prec}$. Moreover, discriminator $D$ is applied to further enhance the realism of generated frames $\hat{y_s}$.
	\begin{equation}
		\mathcal{L}_{\mathrm{adv}}=\mathbb{E}_{y_s}[\log D(y_s)]+\mathbb{E}_{\mathcal{I},\mathcal{I}_s,\beta}[\log (1-D(\hat{y_s}))]
	\end{equation}

	\section{Experiments}
	\subsection{Experimental Settings}
	\noindent \textbf{Datasets} For the Style-E stage, we leverage MEAD dataset~\cite{wang2020mead} with the synthetically generated textual descriptions for emotion styles. MEAD contains videos and audios pairs performed by 60 actors in 8 emotions. Due to the limited size of MEAD, we enhance the one-shot talking motion generation performance by borrowing the pretrained audio encoder from SadTalker~\cite{zhang2022sadtalker}. For the Style-A stage, we additionally utilize another audio-visual dataset HDTF~\cite{zhang2021flow}, which consists of talking videos from more than 300 speakers. To obtain the art style reference, we use various art datasets~\cite{Huo_Gao_Shi_Yin_2017, Huo_Li_Shi_Gao_Yin_2017}. As for the ground truth of Style-A, we employ VToonify~\cite{yang2022vtoonify} to stylize videos in MEAD and HDTF with randomly selected art style $\mathcal{I}_s$.

	\noindent \textbf{Comparison Setting.}
	To the best of our knowledge, there is currently no existing work that can generate high-resolution audio-driven talking face videos with both emotion style and art style from a real image. To provide a comprehensive comparison, we use VToonify to transfer the art style of the input identity image into the desired one. Then, we pass the stylized images through several state-of-the-art (SOTA) talking face generation methods to achieve the same task as our proposed method. The comparing methods include MakeItTalk~\cite{zhou2020makelttalk}, Wav2Lip~\cite{prajwal2020lip}, Audio2Head~\cite{wang2021audio2head}, PC-AVS~\cite{zhou2021pose}, AVCT~\cite{wang2022one}, EAMM~\cite{ji2022eamm} and StyleTalk~\cite{ma2023styletalk}, where only the latter two methods support talking head generation with emotion style. We assess the results using evaluation metrics including SSIM~\cite{ZhouWang2004ImageQA}, FID~\cite{Heusel_Ramsauer_Unterthiner_Nessler_Hochreiter_2017} and PSNR for image generation quality, M-LMD~\cite{chen2019hierarchical} for accuracy evaluation of lip movement, F-LMD~\cite{chen2019hierarchical} for emotion style evaluation. In addition, we calculate $\text{Sync}_\text{conf}$~\cite{JoonSonChung2016OutOT} to measure the synchronization of lip motion with input audio.

	\subsection{Experimental Results}
	\noindent \textbf{Quantitative Results.}
	Table~\ref{tab:quantitative} reports the quantitative results of our method and other SOTA methods. For emotional talking face generation methods (i.e., EAMM and StyleTalk), we additionally provide the emotion videos as emotion resources,  while for our Style$^2$Talker, we use emotion texts as input. Besides, we synthesize high-resolution (1024$\times$1024) talking face videos with both emotion style and art style. As observed, our method outperforms other methods in most of the evaluation metrics on both MEAD and HDTF datasets and achieves a suboptimal score of $\text{Sync}_\text{conf}$, ranking second only to Wav2Lip and AVCT on the MEAD and HDTF datasets, respectively. We argue that this is primarily due to Wav2Lip being trained with the assistance of a SyncNet discriminator, and AVCT receiving an additional phonemes input to enhance the audio-visual correlation. Nevertheless, our lowest M-LMD scores on both datasets demonstrate the satisfactory synchronization between audio and the lip shapes generated by our method.

	\noindent \textbf{Qualitative Results.} 
	Figure~\ref{fig:compare} presents the input for comparison methods marked by `$\#$', the input for ours marked by `*' and the qualitative results. Our Style$^2$Talker achieves the best lip synchronization and both emotion style and art style transfer in high resolution directly from the real face. Specifically, MakeItTalk struggles to generate accurate lip motions and Wav2Lip suffers from blurriness in the mouth region. Audio2Head and PC-AVS encounter issues with identity information loss. Despite the progress achieved by AVCT, it neglects the emotion style. While EAMM and StyleTalk take an emotion source video for emotion style reference which provides more concrete style information than textual description used in our method~\cite{ma2023talkclip}, we perform more expressive emotion style than EAMM, which appears server facial deformation as circled in the red boxes. In comparison to StyleTalk, we achieve competitive emotion style performance. However, whe n silent, their mouths remain open (pointed by red boxes), which leads to unnatural lips and noticeable artifacts (pointed by yellow boxes).

	\begin{figure}[t]
		\centering
		\begin{subfigure}{0.32\linewidth}
			\includegraphics[width=1\linewidth]{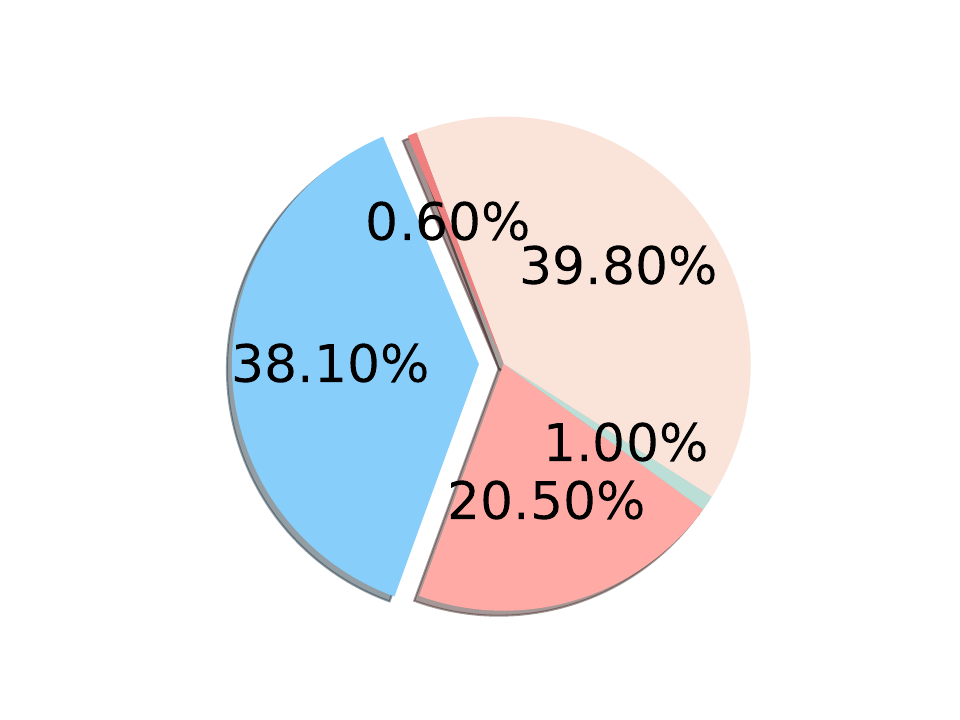}
			\caption{Emotion Style}
			\label{fig:userstudy1}
		\end{subfigure}
		\begin{subfigure}{0.32\linewidth}
			\includegraphics[width=1\linewidth]{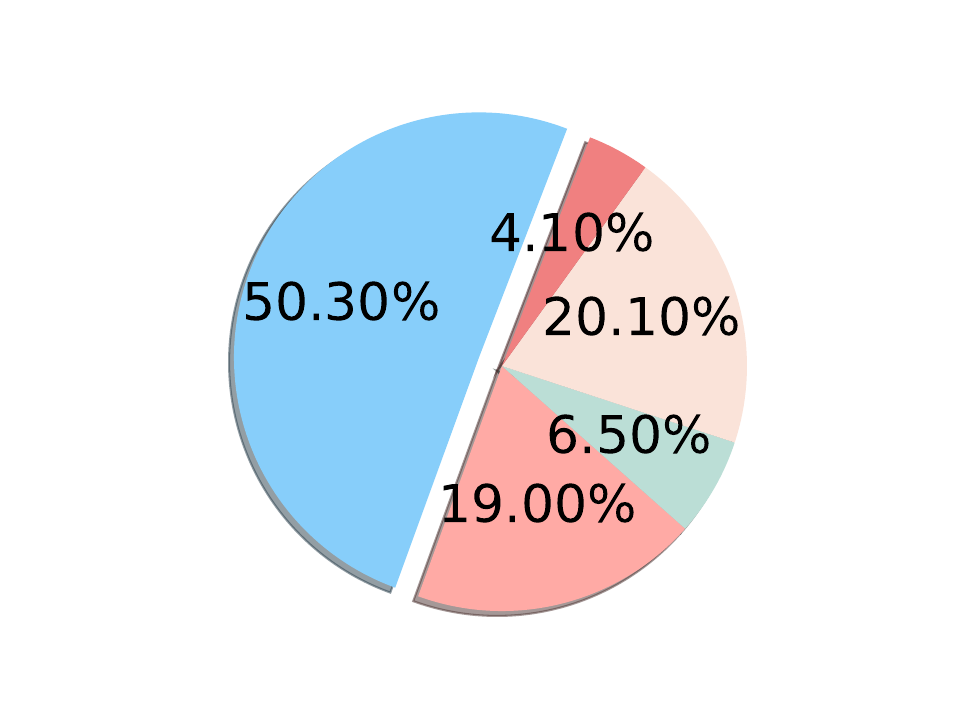}
			\caption{Art Style}
			\label{fig:userstudy2}
		\end{subfigure}
		\begin{subfigure}{0.32\linewidth}
			\includegraphics[width=1\linewidth]{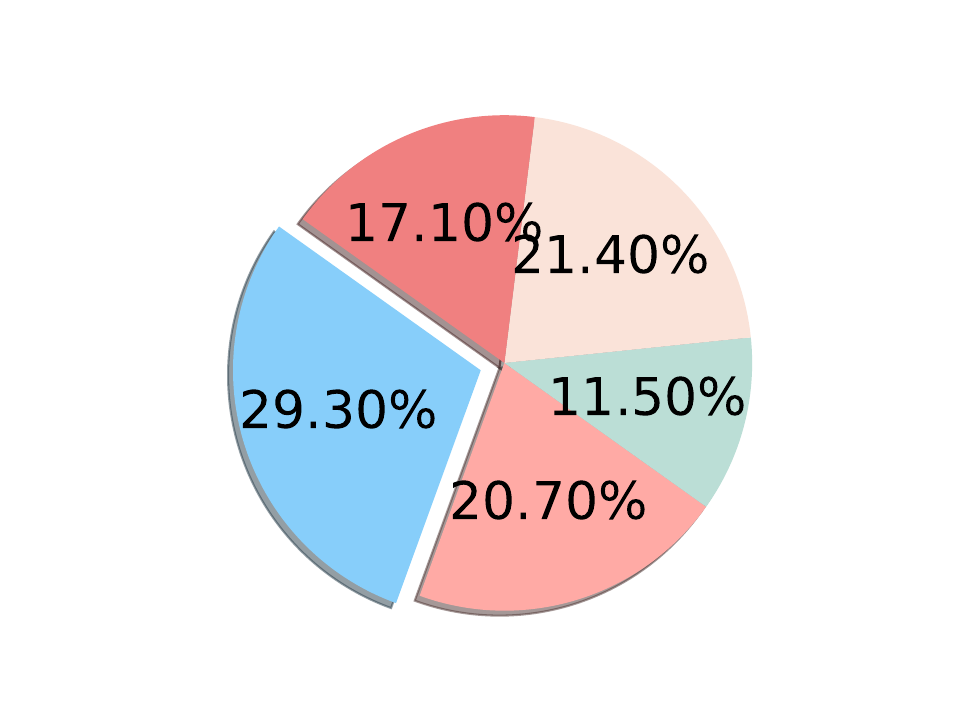}
			\caption{Lip Sync}
			\label{fig:userstudy3}
		\end{subfigure}
		
		\begin{subfigure}{1\linewidth}
			\includegraphics[width=1\linewidth]{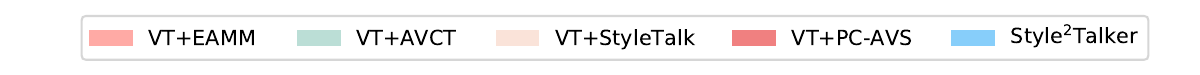}
		\end{subfigure}
		\caption{User study results.}
		\label{fig:userstudy}
	\end{figure}
	
	\noindent \textbf{User Study.}
	We conduct user studies to compare our method with SOTAs in terms of human likeness. Specifically, we invite 20 participants (10 males + 10 females) and each participant is presented with 5 videos generated by 5 different methods (2 SOTA methods each without/with emotion style and Style$^2$Talker) per iteration. Participants are required to choose from among them the one in which they believe the \textbf{emotion style} best matched the provided text/the \textbf{art style} best matched the given picture/the \textbf{lip motion} best matched the audio, and such a process is repeated 20 iterations. The results, as depicted in Figure~\ref{fig:userstudy}, indicate that our method received the most preferences for art style and lip synchronization aspects and competitive likeness to StyleTalk in emotion style, which inputs a more informative emotional video.

	\begin{table}
		\centering
		\resizebox{\linewidth}{!}{
			\begin{tabular}{@{}l|cccc@{}}
				\toprule
				Method/Score & SSIM $\uparrow$ &FID $\downarrow$ & M/F-LMD $\downarrow$& $\text{Sync}_\text{conf}$ $\uparrow$\\
				\midrule
				w/o emo style & 0.760 & \underline{31.655} &\underline{3.269}/ 2.913 & \textbf{3.762}\\
				w/o art style & 0.743 &58.119 & \textbf{ 3.132} / 2.897  & 2.477\\
				w/o dif model & \underline{0.789} & 24.685 & 4.339 / 3.712& 2.521\\
				w/o skip con. & 0.763 & 42.105 & 3.868 / 3.082& 2.567\\
				w/o $R$ & 0.754 & 35.530 & 3.557 / \underline{2.882}& 2.749\\
				\midrule
				\textbf{Full model} & \textbf{0.795} & \textbf{23.207}& 3.317 / \textbf{2.696} & \underline{2.847}\\
				\bottomrule
			\end{tabular}
		}
		\caption{Results for ablation study.}
		\label{tab:ablation}
	\end{table}

	\begin{figure}[t]
		\centering
		\includegraphics[width=0.95\linewidth]{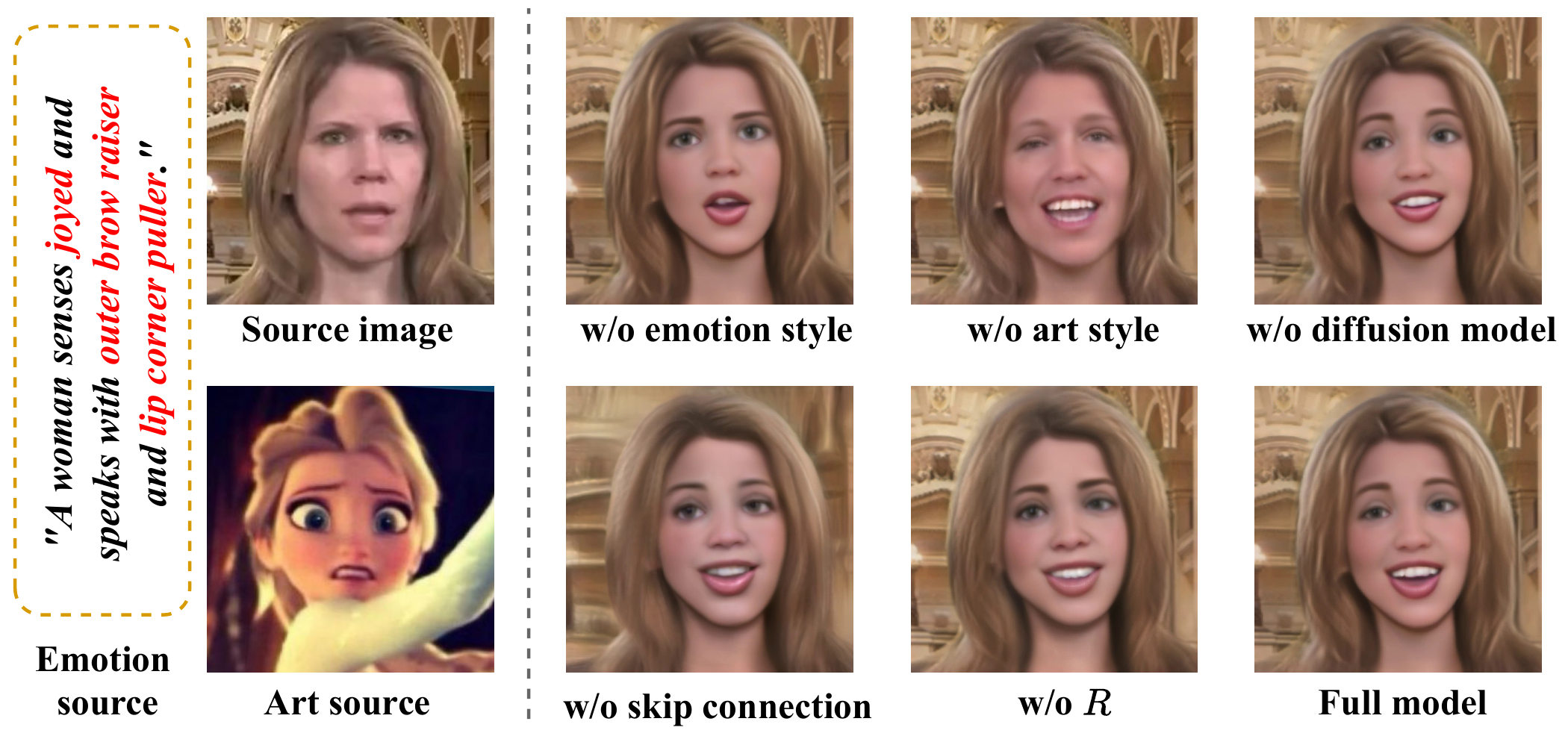}
		\caption{Visualization Results of ablation study.}
		\label{fig:ablation}
	\end{figure}
	\noindent \textbf{Ablation Study.}
	We further conduct ablation experiments on MEAD dataset to assess the effectiveness of each introduced component. The qualitative and quantitative results are presented in Figure~\ref{fig:ablation} and Table~\ref{tab:ablation}. In general, the experiment settings and corresponding analyses are summarized as: \textbf{(a)} w/o emotion style: we exclude the input of emotion source text $\mathcal{T}$ and the text encoder of CLIP. As a result, the expressions in the generated videos are consistent with the identity image, and the F-LMD score decreases accordingly. This indicates that our method is capable of comprehending and incorporating the emotion style conveyed in the text description. \textbf{(b)} w/o art style: we build our model upon the vanilla StyleGAN instead of DualStyleGAN by removing the art path and ModRes block. In this case, the art style remains unchanged, but the FID score significantly increases due to the different art style from the ground truth.  This confirms that the art path and ModRes block are crucial for maintaining consistent and visually appealing art style transfer. \textbf{(c)} w/o diffusion model: we replace the diffusion model with an conditional GAN~\cite{mirza2014conditional} in the Style-E stage, which results in worse lip synchronization. This observation illustrates that the diffusion model achieves superior conditional generation performance. \textbf{(d)} w/o skip connection and \textbf{(e)} w/o $R$: we remove middle-level skip connection and refinement network $R$, respectively. As shown in Figure~\ref{fig:ablation}, the detail disparity between our full model and the source identity image is smaller than the disparity of \textbf{(d)} w/o skip connection. Furthermore, our full model exhibits higher fidelity with the source image than \textbf{(e)} w/o $R$, demonstrating that the skip connections and $R$ effectively enhance the image quality and maintain better visual coherence.

	\section{Conclusion}
	In this paper, we present Style$^2$Talker, a novel system that generates high-resolution emotionally and artistically stylized talking face videos by incorporating corresponding style prompts. Leveraging a labor-free text annotation pipeline based on large-scale pretrained models, we obtain textual descriptions for emotion style learning from text inputs. We aspire for our attempt to inspire further in-depth research, employing outstanding large-scale pretrained models for more practical and captivating explorations. To infuse emotion style into the 3D motion coefficients, we devise an efficient diffusion model with multiple encoders, ensuring the generation of realistic and expressive facial expressions. We incorporate a motion-driven module and an additional art style path into the StyleGAN architecture, enabling coefficient-driven video generation with desired emotion and art styles. To further enhance the visual quality and eliminate artifacts, we employ a content encoder and refinement network. Qualitative and quantitative experiments demonstrate that our method can generate more stylized animation results compared with state-of-the-art methods. 
	
	\section{Acknowledgments}
	This work was supported by National Natural Science Foundation of China
	(NSFC, NO. 62102255) and Shanghai Municipal Science and
	Technology Major Project (No. 2021SHZDZX0102). We would like to thank Xinya Ji, Yifeng Ma and Zhiyao Sun for their generous help.

	\bibliography{style2talker}
	
\end{document}